# An Object-Oriented MDP Representation for Robotic planning

**Aasheesh Singh, Department of Electronics & Communication, Delhi Technological University**
(asjsingh97@gmail.com)

*Abstract*—**This paper aims to implement Object-Oriented Markov Decision Process (OO-MDPs) for goal planning and navigation of robot in an indoor environment. We use the OO-MDP representation of the environment which is a natural way of modeling the environment based on objects and their interactions. The paper aims to extend the well known Taxi domain example which has been tested on grid world environment to robotics domain with larger state-spaces. For the purpose of this project we have created simulation of the environment and robot in ROS (Robot operating system) with Gazebo and Rviz as visualization tools. The mobile robot uses a 2D LIDAR module to perform SLAM in the unknown environment. The goal of this project is to be able to make an autonomous agent capable of performing planning and navigation in an indoor environment to deliver boxes (*passengers* in Taxi domain) placed at random locations to a particular location *(warehouse)*. The approach can be extended to a wide variety of mobile and manipulative robots**

## 1. INTRODUCTION

Markov Decision process is a discrete time stochastic control process. At each time step, the agent is in some state **s,** and takes an action **a** which takes it to its next state **s'** giving the agent a reward **$R_a(s,s')$.** The core problem of MDPs is to find a "policy" for the decision maker: a function $\pi$ that the decision maker will choose when in state **s.** Algorithms for Reinforcement learning in MDP environments suffer from what is known as the curse of dimensionality: an exponential explosion in the total number of states as a function of the number of state variables. Learning in environments with extremely large state spaces is challenging if not infeasible without some form of generalization.

The OO-MDP representation starts from attributes that can be directly perceived by the agent, rather than predicates introduced by the designer, this similar kind of formalism has been used in Relational MDPs (RMDPs), introduced by Guestrin et al. (2003) in the context of planning. The *Taxi* domain example has been successfully tested with a videogame "*Pitfall*" in the paper "*An Object-Oriented Representation for Efficient Reinforcement Learning*" by Deuk, Littman & Cohen. We aim to extend this approach to mobile robots in an indoor grid like environment.

## 2. Notation

We use a standard Markov Decision Process (MDP) notation throughout this paper (Puterman, 1994). A finite MDP M is a five tuple <S,A,T,R,γ>. We use $T(s'|s,a)$ to denote the transition probability of state s' given state– action pair(s,a) and $R(s,a)$ to denote the expected reward value. A deterministic MDP is one in which there is a single next state s' for every given state s and action a; that is, $\forall s \in S$, $a \in A$, $\exists s' \in S$: $T(s'|s,a)=1$.

## 3. Environment Representation

We will use the Taxi domain, defined by Dietterich (2000), as an example to introduce our formalism. Taxi is a grid world domain (see Figure 1.a), where a taxi has the task of picking up a passenger in one of a pre-designated set of locations (identified in the figure by the letters Y, G, R, B) and dropping it off at a goal destination, also one of the pre-designed locations. The set of actions for the taxi are North, South, East, West, PICKUP and DROPOFF. Walls in the grid limit the taxi's movements. Fig. 1.b shows the simulated indoor environment in Gazebo consisting of walls, obstacles and cylindrical boxes to be picked (*passengers*).

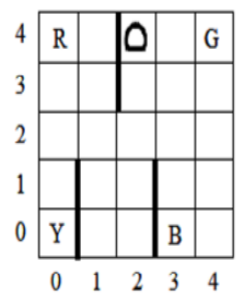

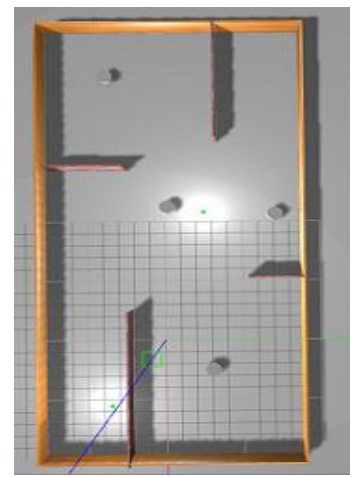

| S.no | Class name | Attributes of class | No. of objects |
|------|-----------|---------------------|----------------|
| 1. | Agent | Odometry data | 1 |
| 2. | Cylindrical Boxes(passenger) | Odometry, Boolean variable in_bot | Variable |
| 3. | Walls/Obstacles | Odometry coordinates | Variable |
| 4. | Destination | Odometry coordinates | 1 |

The OO-MDP representation of the environment consisting of its objects is shown in the tabular form. The class Passenger has an additional attribute in.bot which returns true, when the object has been picked.

When two objects interact in some way, they define a **relation** between them. A combination of the relation established, plus the internal states of the two objects, determines an effect—a change in value of one or multiple attributes in either or both interacting objects. This is a *significant change in property* from classical MDPs which consider walls as a property of the specific location of the grid whereas in OO-MDP , the interaction(hence the relation) is same irrespective of the location. For our bot representation, we will define 5 relations: $touch_N(o1,o2)$, $touch_S(o1,o2)$, $touch_E(o1,o2)$, $touch_W(o1,o2)$ and $on(o1,o2)$, which define whether an object $o2 \in C_j$ is exactly one cell North, South, East or West of an object $o1 \in C_i$, or if both objects are overlapping (same x, y coordinates).

## 4. Data structure and definitions:

- **T** is the union of all terms t that will be involved in the conditions that determine the transition dynamics of the environment described by the OO-MDP, plus their negations ¬t, with |T|=2n.
- **Cond(s)** function returns all the terms that are true in state s.
- An **effect E** is change of attributes required to cause a transition from state s to s' with action a.
- For comparing two conditions c1 & c2 we have the commutative operator defined as :

| $c_1$ | $c_2$ | $c_1 \oplus c_2$ |
|-------|-------|------------------|
| 0 | 0 | 0 |
| 1 | 1 | 1 |
| 0 | 1 | * |
| 0/1 | * | * |

- For any states **s** and **s'** and attribute **att**, the function **eff_att(s,s')** returns one effect of each type that would transform attribute att in s into its value in s'.
- A prediction p is a pair (**p.model, p.effect**), where p.model is a condition that represents the set of terms that need to be true for p.effect to occur.
- If for an action a, s=s' then that condition is called *failure condition,* denoted by **F_A**.

- The relations together with the attributes define the state of the system, here we will have a 7 bit array describing the state of the system as: Suppose the bot is in the position (2,4) of the grid world defined in fig. 1then **cond(s)** returns:

{$touch_N$(bot,wall), ¬ $touch_S$(bot,wall),
¬ $touch_E$(bot,wall), $touch_W$(bot,wall),
¬on(bot,passenger),¬on(bot,destination),
passenger.in_bot = T }
forming a **7 bit array= 1001001**.

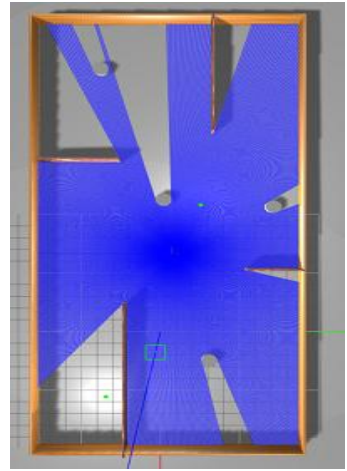

Fig. 3: Overhead view of the Laser graph when the bot is placed in the middle of the room. Passenger boxes & obstacle walls deflect the laser beam while it passes through the rest of the area(shown in blue). From here, we can form the touch(o1,02) relations between objects.

## 5. Mapping and localization

The robot uses **Adaptive Monte-Carlo localization** to localize itself in the environments given the map of the environment. In ROS the "gmapping" package generates the map of the environment given the laser scan data, the transform frame and odometry topics. The approach derives from earlier work on Markov localization which represents the robot's belief by a probability distribution over possible positions, and use Bayes rule and convolution to update the belief whenever the robot senses or moves. The idea of probabilistic state estimation goes back to Kalman filters which use multivariate Gaussians to represent the robot's belief. MCL uses fast sampling techniques to represent the robot's belief. When the robot moves or senses, importance resampling is applied to estimate the posterior distribution. An adaptive sampling scheme which determines the number of samples on-the-fly, is employed to trade-off computation and accuracy.

### A. Advantages of using Monte carlo localization over Kalman filter and Markov localization:

- In contrast to existing Kalman filtering based techniques, it is able to represent multi-modal distributions and thus can globally localize a robot.
- It drastically reduces the amount of memory required compared to grid-based Markov localization and can integrate measurements at a considerably higher frequency.
- It is more accurate than Markov localization with a fixed cell size, as the state represented in the samples is not discretized.

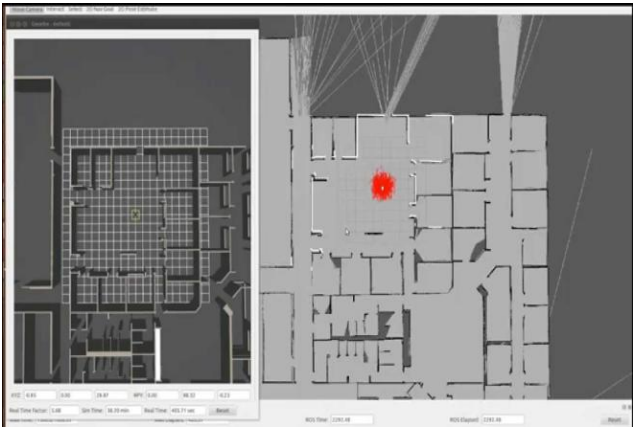

Fig. 3 AMCL simulation on a bot in a maze like environment. The red cloud around the bot shows the robot's belief/uncertainty in its pose while localization.

Amcl takes in a laser-based map, laser scans, and transform messages, and outputs pose estimates. On startup, amcl initializes its particle filter according to the parameters provided.

## 6. Discovering the optimal policy/Optimisation:

### A. General transition dynamics:

For each state-action pair function **predictTransition(s,a)** returns a predicted next state s' according to *p.model*. The agent tries to discover the optimal policy $\pi$ by running repetitive episodes of the same task to convergence.

The two main routines of the algorithm are predictTransition (Algorithm 1), which pre-dicts the next state given a current state and action based on the current model, and **addExperience** (Algorithm 2), which learns a model of the OO-MDP. If predictTransition is not able to predict a next state with accuracy, it returns $s_{max}$.

We state the algorithm for the **predictTransition(s,a)** as follows:

**Input:** (s,a)  **Output:** predicted state s'

**1: if** $\exists c \in F_a$ s.t. $c \models cond(s)$ **then**

   // C is a known failure condition in the environment( **eg. Collision with walls and obstacles**)

   Return s

**2: else**

**3:** **for all** attributes att $\in U_{c \in C}$ Att(c) **do**

**4:** E $\leftarrow \emptyset$

**5:** **if** $\exists p \in pred(a, att, type)$ s.t. $p.model \models cond(s)_S$ **then**

**6:** Add $p.effect$ to $E$

**7:** **end if**

**8.** **end for**

**9:** **if** $E = \emptyset \vee \exists e_i, e_j \in E$ s.t. $e_i$ and $e_j$ are incompatible **then**

**10:** Return $s_{max}$

**11:** **else**

**12:** // Set $E$ contains all the individual operations that need to be applied to attributes in $s$ in order to convert it to s'.

**13:** $s \leftarrow$ apply $E$ to $s$

**14:** Return s

**15:** **end if**

**16: end if**

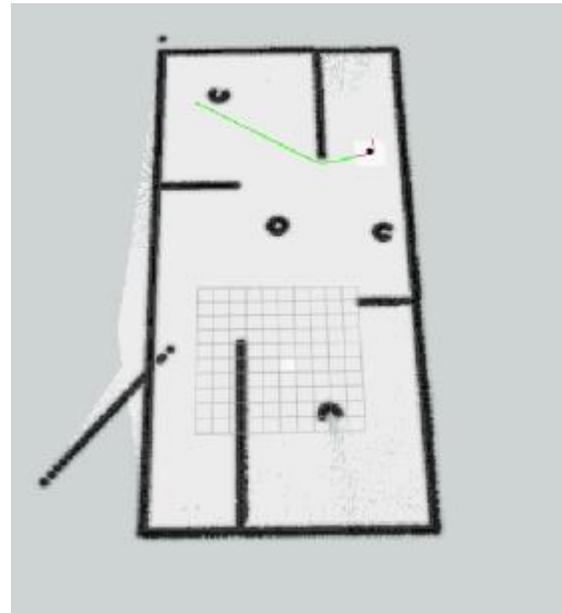

Fig. 4: Rviz simulated map of the grid like indoor environment. Gmapping node returns the map of the environment using the data published on the /sensor_mgs/LaserScan topic by the 2D LIDAR module. The green line indicates the predicted path(next state) of the bot as it advances.

We will now state a theorem to state the worst case bound i.e. the maximum no. of steps required to attain optimal policy so that we can state our next algorithm on **addExperience(s,a,s',k)**

**Theorem:** *The transition model for a given action a, at-tribute att and effect type type in a deterministic OO-MDP is KWIK-learnable with a upper bound of $O(nk + k + 1)$, where n is the* *number of terms in a condition and k is the maximum number of effects per action–attribute.*

---

**Algorithm 2** `addExperience(s,a,s',k)` method

---

0: **Inputs:** an observation $< s, a, s >, k$; the maximum number of different effects possible for any action, attribute and effect type.

1: **if** $s = s$ **then**

2:      // Found a failure condition for action $a$, update $F_a$

3:      Remove all $c \in F_a$ s.t. $cond(s) \models c$.

4:      $F_a \leftarrow F_a \cup \{cond(s)\}$

5: **else**

6:      **for all** attributes $att \in \cup_{c \in C} Att(c)$ **do**

7:          **for all** $e \in eff_{att}(s, s)$ **do**

8:          Find a prediction $p \in pred(a, att, e.type)$ such that $p.\text{effect} = e$

9:          **if** $\exists p$ **then**

10:            // We already have a (condition, effect) prediction for current $a$, $att$, and $type$. Update condition and verify that there are no over-laps.

11:            $p.\text{model} \leftarrow p.\text{model} \oplus cond(s)_S$.

12:            **if** $\exists c \in (pred(a, att, e.type) \setminus p).\text{models}$ s.t. $p.\text{model} \models c$ **then**

13:              // Conditions overlap, violating an assumption, meaning it is not the right *type* of effect for this action and attribute.

14:              Remove $pred(a, att, e.type)$ from $P$

15:            **end if**

16:          **else**

17:            // We observed an effect for which we had no prediction. If its condition does not over-lap an existing condition, then add this new prediction.

18:            **if** $\exists c \in pred(a, att, e.type).\text{models}$ s.t. $cond(s) \models c \lor c = cond(s)$ **then**

19:              Remove $pred(a, att, e.type)$ from $P$

20:            **else**

21:              Add $(cond(s), e)$ to $pred(a, att, e.type)$.

22:              // Verify that there aren't more than $k$ predictions for this action, attribute and type.

23:              **if** $|pred(a, att, e.type)| > k$ **then**

24:                 Remove $pred(a, att, e.type)$ from $P$

25:              **end if**

26:            **end if**

27:          **end if**

28:          **end for**

29:      **end for**

30: **end if**

## 7. Realization in Robot operating system(ROS)

We form a grid world like structure mirroring the Taxi-domain example by dividing our environment with grid size= footprint of the robot. The physical meaning of the various parameters as described in the taxi domain can be represented as:

- S= current x,y coordinate frames returned by odometry data and determining the pose using AMCL algorithm upto certain belief.
- S'= next estimated state for goal-planning

- The laser scan data returns the distance from a obstacle nearby which is divided into grid units and the therefore the relation touch(agent,wall) is formed which tells the bot of next possible actions. For eg: If $touch_N$(agent,wall) is true the bot cannot give velocity commands in the north/forward direction as it would lead to a failure state or collision.

- The effect type E gives all the possible change in attributes(x,y coordinates) that has to take place according to the policy.

- After every episode/simulation the robot returns data collected in the function pred(a,att,e.type) which is updated over the next episode and hence optimal policy to discover the shortest possible path to achieve the goal is discovered over various iterations. (Algorithm 2)

## 8. Conclusion and Future work:

MDP like environment suffer from what is known as the curse of dimensionality due to the exponential number of state variables as state-spaces increase. OO-MDP greatly simplifies this problem by reducing state-spaces by making MDP properties(location of walls, obstacles) as objects. But still the approach is currently valid for static transition dynamics as for stochastic environments a more complex learning algorithm would be needed to learn transitions effectively in the face of noise. We aim for extending this approach to much dynamic environments and hence a rather more effective learning algorithm for it.

## 9. Acknowledgment

I would like to thank Dr. Balaraman Ravindran, Dept. of Computer Science, IIT Madras under whose guidance i got the opportunity to do my first research project on Reinforcement learning. Also i would like to thank my mentors Priyatosh Mishra (Phd.) and Mohammed Hisham at RISE Lab, IIT Madras for guiding me at every step of the project and learning experience.